# Measuring Sociality in Driving Interaction

Xiaocong Zhao, Jian Sun, Meng Wang, *Member, IEEE*

*Abstract*— Interacting with other human road users is one of the most challenging tasks for autonomous vehicles. For congruent driving behaviors, it is essential to recognize and comprehend sociality, encompassing both implicit social norms and individualized social preferences of human drivers. To understand and quantify the complex sociality in driving interactions, we propose a Virtual-Game-based Interaction Model (VGIM) that is parameterized by a social preference measurement, Interaction Preference Value (IPV). The IPV is designed to capture the driver's relative inclination towards individual rewards over group rewards. A method for identifying IPV from observed driving trajectory is also developed, with which we assessed human drivers' IPV using driving data recorded in a typical interactive driving scenario, the unprotected left turn. Our findings reveal that (1) human drivers exhibit particular social preference patterns while undertaking specific tasks, such as turning left or proceeding straight; (2) competitive actions could be strategically conducted by human drivers in order to coordinate with others. Finally, we discuss the potential of learning sociality-aware navigation from human demonstrations by incorporating a rule-based humanlike IPV expressing strategy into VGIM and optimization-based motion planners. Simulation experiments demonstrate that (1) IPV identification improves the motion prediction performance in interactive driving scenarios and (2) the dynamic IPV expressing strategy extracted from human driving data makes it possible to reproduce humanlike coordination patterns in the driving interaction.

## I. Introduction

WITH the speeding deployment of autonomous vehicles (AVs) both in the testing and commercialization phases [1], interactions with other road users are increasingly acknowledged to hinder AV development [2], [3]. It is reported that the failure in driving interactions *where road users' navigations are necessarily affected by others' intended future actions* is one of the key contributors to AV-related traffic accidents [4], [5]. Aside from traffic accidents which bring direct and observable consequences, AV's failure in driving interactions with manually driven vehicles can also impede traffic efficiency [6], [7], fatigue surrounding human drivers [3], and even lead to road rage [1], [8].

The reason for failed AV-involved driving interactions is usually not the violation of the traffic law but the insufficient

This work was supported in part by National Natural Science Foundation of China 52125208, the Shanghai Municipal Science and Technology Major Project 2021SHZDZX0100, the Fundamental Research Funds for the Central Universities 2022-5-ZD-02, and in part by Zhejiang Lab Open Research Project NO.2021NL0AB02. *(Corresponding author: Jian Sun.)*

Xiaocong Zhao and Jian Sun are with Department of Traffic Engineering & Key Laboratory of Road and Traffic Engineering, Ministry of Education, Tongji University, Shanghai, China (e-mail: zhaoxc@tongji.edu.cn; sunjian@tongji.edu.cn ).

Meng Wang is with Chair of Traffic Process Automation, Technische Universität Dresden, Dresden, Germany (e-mail: meng.wang@tu-dresden.de).

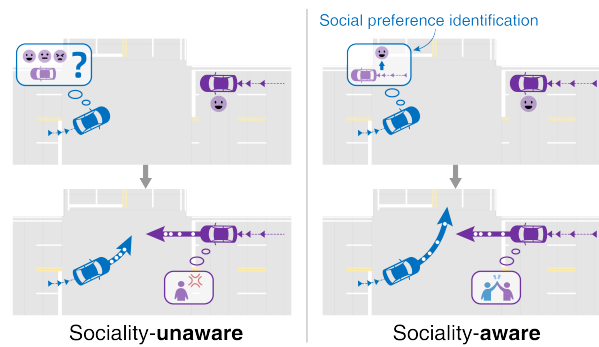

Fig. 1. A motivating example of VGIM. **Left**: The blue car is unaware of the sociality and does not take the right of way given by the purple car, resulting in a longer standstill for both two cars. **Right**: The driving interaction evolves smoothly as the blue car detects the purple car's cooperative social cue and speeds up in order to echo the purple car's willingness of giving the right of way.

understanding of traffic social rituals. In fact, AVs are good at navigating legally due to their programmable nature [3]. However, written traffic laws only structure the basic traffic order and still leave ambiguous situations where the order of access is not clarified explicitly. To deal with navigation in these situations, implicit rules have been developed and are usually applied in human driving interactions. As shown in Fig. 1, although traffic laws have suggested the passing order in most shared spaces, human drivers with a higher priority (purple car) may sometimes give the right of way to their interacting counterparts (blue car) due to courtesy and respect to others [9]. In the left case of Fig. 1, the blue car, without the awareness of social preference, cannot detect and fails to echo the cooperativeness of the purple car, resulting in a long standstill. In the right case, the blue car's snatching the right of way seems to be not fully in line with the official traffic laws but does fulfill the human expectation. Systematic implicit rules have been established as social rituals in the human driver community to guide *socially preferable* driving behaviors, compensating for the silence of written laws. To get into the same communication channel and congruently interact with human drivers, AVs also have to understand and harness these implicit rules in driving interactions [7], [10].

While the general concept of the driving interaction is intuitive, the mathematical description and modeling of interactions are still open and challenging questions. Two main issues are to be addressed. First, the interplay among agents makes unilaterally modeling a single agent's behaviors almost impossible. Actions in interactive scenarios are influenced by the surrounding agents and conversely play a role in the evolvement of the environmental dynamics. Therefore, modeling the driving interaction is a task not of describing how one acts on the existence of others but of revealing the inter-agent action dependence. Second, with at least two drivers involved, the driving interaction possesses the nature of human socializing. As is widely reported, human drivers exchange implicit social cues with surrounding vehicles through communicative behaviors [1], [11]–[13], for example, asserting the right of way by maintaining the approaching speed. The social preference in these communitive behaviors greatly influences the interaction evolvement [7], [10]. Therefore, to model sociality-featured driving interactions, it is essential to take into account the social preferences of interacting agents. Key points of embedding social preference in AV design may include (1) decoding the social preference expression from observable information and (2) learning a human-like strategy for expressing social preferences.

Serving the purpose of instructing congruent AV navigation in human-involved driving interactions, the objective of this study is to provide an approach for measuring and analyzing the social preference in human driving interactions. We propose a driving interaction model, the Virtual-Game-based Interaction Model (VGIM), that employs game theory to capture inter-agent action dependence. In VGIM, the Interaction Preference Value (IPV) is designed to quantify individualized social preference. Based on the VGIM, we analyze the social preference of human drivers with real-world driving data and present insights for designing socially preferable AV based on the results. The main contribution of this work is three-fold:

- Proposing a generalizable indicator of social preference (i.e., IPV) that can be estimated from observed driving trajectories.
- Developing a game-theoretic model to describe the social-preference-aware driving interaction between heterogeneous agents.
- Presenting a method for socially compliant AV design by learning the IPV expressing strategy from the human demonstration in driving interactions.

The rest of this paper is arranged as follows: Firstly, previous works related to driving interaction modeling and social preference analysis are reviewed in §2. Then, we formulate the VGIM in §3 to mathematically describe human drivers' motivation, action dependence, and information-gathering process in the driving interaction. In §4, we analyze the social preference in human driving interactions using VGIM. Based on the analysis results, discussion and simulation experiments are conducted in §5 to bring the human strategy in the driving interaction into light. Finally, we conclude our work in §6.

## II. RELATED WORKS

Being critical to traffic safety, driving interaction has received attention in a wide range of fields, including automation, sociology, and linguistics [9]. The main focus of this study is to model the driving interaction and quantitatively analyze the social preference in human driving. Therefore, we narrow our review scope down to the above two topics.

### A. Driving interaction modeling

*Modeling interactive agents*. The bloom of modeling interactive agents largely comes with the desire to design AVs that are capable of navigating in human-involved interactive driving scenarios. To verify AV's performance in interacting with others, one theoretically needs background vehicles that have the interactive driving ability. However, calling for the interactive driving ability to design interactive driving strategies could then become a "chicken and egg" problem [14]. A common solution is to assemble a human-driven vehicle (HV) model with certain capabilities to respond to the AV under test. According to *level-k* theory, decision-makers have different levels of reasoning ability [15]. A level-0 agent does not cooperate with others and behaves reflexively. For k>0, a level-(k) agent assumes all the other agents are level-(k-1) and acts accordingly. Experimental studies in economics demonstrate that most people are no higher than level-2 decision-makers [16], [17]. Following the *level-k* theory, one could construct reasonably simplified background traffic where the HV model's level of reasoning varies from *reactive* (respond directly to the real-time state) [18], to *proactive* (re-plan oneself based on the prediction of others' motion) [19], [20], or even more sophisticated [21], [22]. Some studies also introduce random parameters to allow individualization in modeled HVs within this diagram [14]. Serving the purpose of testing AV, interactive agent models were usually only tuned for better reproducing the behaviors of a single HV. However, putting together a group of such interactive agents does not ensure a humanlike driving community from which coordination patterns could emerge.

*Modeling interaction events*. Modeling interactions at an event level needs to further consider the action dependence among agents, namely, how interacting agents affect one another. Game theory is a typical and popular method for this purpose due to its nature of modeling conflict and cooperation between rational decision-makers [23]. We observed two main branches of applying game-theoretic models in dealing with driving interaction modeling. The Stackelberg game employs a leader-follower structure to explain scenarios where interacting agents have distinct right of way, for example, ramp merging [24] and lane changing [25], [26]. Nash game was applied for more general driving interaction events. Without manually settled leader-follower relations, the Nash game is suitable for modeling multi-agent driving interactions but also takes its toll on intractable computational complexity [27]. In recent years, state-of-the-art technologies for efficiently computing games have been proposed in the robot community, making the game-theoretic model a promising trend in driving interaction modeling [28].

### B. Social preference representation

Social preference is a kind of individualized property that influences human behaviors in group events [29]. People with different social preferences diversely weigh ego and others' rewards when involved in group events [30], including driving in a shared traffic space [9], [31]. With the awareness of social preference in driving interactions, two branches of studies are developed to design socially congruent AV. The first branch believes that human-friendly automation is the solution. For instance, Bahram et al. [20] argue that AVs should be cooperative when interacting with HVs. Being always kind to HVs seems not to harm, but concerns have been raised, worrying that human drivers may gradually develop a tendency to bully cooperative AVs in the long term [6]. The second branch holds the view that AVs should behave in a human-like way to gain equal positions when interacting with HVs. Ren et al. [8] integrated social gracefulness into AV planners and allowed AV to learn sophisticated human driving behaviors. However, as was designed for measuring AV's reaction to HVs, social gracefulness's adaptability to analyzing human behaviors remains unproven. Supported by mature psychological studies, Schwarting et al. [32] conducted another typical research by introducing a psychological term, the social value orientation (SVO), into driving interaction modeling to capture the social preference of interacting agents. Schwarting's work proved that by taking social preference into account, the game-theoretic model could better explain and reproduce human driving behaviors. This finding is one of the foundations of our work.

Though with great efforts made, major knowledge gaps still exist. First, although being aware that driving interaction is a sociality-featured event, we still have little understanding of the human-like way of expressing social preferences, which is essential for AV design. Second, the direct use of psychological indicators like SVO makes it necessary to divide human drivers' utility terms into ego and others' rewards [22], [32], while coordination in some interaction scenarios brings utility terms that are hard to categorize under this setting (e.g., collision avoidance benefits both ego and interacting vehicles). Therefore, technically, we still lack a cross-scenario applicable measurement of social preference for consistent analysis of sociality in human driving interactions. To fill the gaps above, this study constructed a simple-yet-generalizable driving interaction model and provided a quantified framework for analyzing human drivers' sociality patterns in the driving interaction.

## III. PROBLEM DESCRIPTION AND MODELING

This section formulates the driving interaction. For definiteness and without loss of generality, we first give a general description of the main components in the proposed framework. Then, the detailed formulation is provided by concreting the model with a typical real-world driving scenario, the unprotected left turn.

### A. General description

We assume human drivers are utility-maximizing agents in this study. Considering driving interactions are group events that happen among at least two agents, we argue that human drivers' desired rewards could be divided into two kinds, namely, individual and group rewards. The *individual rewards* cover rewards that rely only on ego actions, including travel efficiency, comfort, compliance with traffic rules, etc. Correspondingly, *group rewards* depend not only on ego actions but also on the actions of other interacting agents. For example, due to the limited road resources, one needs to share the road with other road users in some circumstances, like ramp merging in dense traffic, unprotected left-turning, etc., which may lead to potential collisions. As avoiding collisions relies on all involved agents' actions, it brings group rewards. With the criterion of dividing rewards, we define *driving interaction* as *a situation where more than one motorized road user compromises part of its potential individual rewards as a result of taking into account group rewards*. Note that as all interacting agents jointly influence the group rewards, one has to estimate how much effort for group rewards its interacting agents are make, i.e., the interacting agent's social preference, in order to pursue group rewards purposefully.

With the definition of driving interaction above, we can construct a driving interaction model by combining three basic components: (1) Motivations: the specifications of individual and group rewards; (2) Action-dependence: the interplay of interacting agents' behavior choices; (3) Information gathering: the process of gradually generating self-fulfilling belief on the interacting counterpart's social preference.

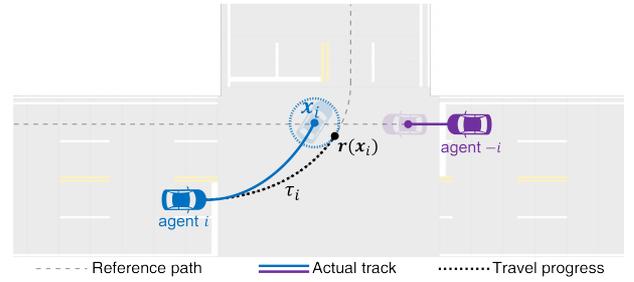

Fig. 2. The unprotected left-turn scenario.

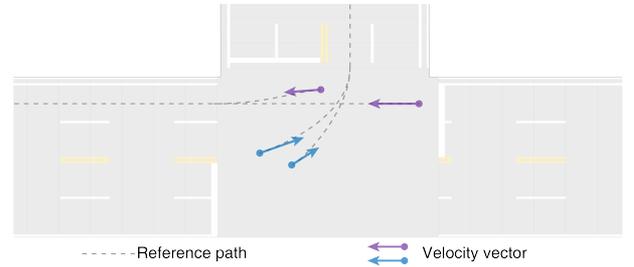

Fig. 3. Dynamic reference path.

### B. Formulation of Virtual-Game-based Interaction Model

In this subsection, we detailed our driving interaction modeling framework in the two-agent unprotected left-turn scenario by exemplifying the reward expression and formulating the action-dependence and the information-gathering process with a game-theoretic model.

#### 1) Motivation

Consider the unprotected left-turn scenario where left-turn (LT) and straight-through (ST) agents traverse across the intersection with conflicts along their future paths (Fig. 2). This problem is mathematically symmetric for both agents as they share the same purpose of getting across the conflict point. We could therefore roll out the formulation from the perspective of either of the two agents. Without loss of generality, we use subscripts $i$ and $-i$ to index the subjective agent and its interacting counterpart, respectively. The following formulations indexed by $i$ are exemplified from the view of the LT agent but applicable for both. We assume that two agents cannot communicate with each other explicitly and only the position $x \in \mathbb{R}^2$ and the heading angle $\psi \in \mathbb{R}$ can be observed.

*Individual reward.* Putting aside the potential collisions between interacting agents which will be discussed later, the individual target for every single agent could be stated as: To plan an $N$-segments trajectory $x_i = x_i^{0 \to N} \in \mathbb{R}^{2 \times (N+1)}$ to cross the intersection as soon as possible. Here, *trajectory* refers to a path coupled with a speed profile that can be followed by a vehicle. The position $x_i^n$ is the end of the $n$-th segment ($n \in [1, N]$) and $x_i^0$ is the starting position. Agents are assumed to travel every segment within a constant time interval $\delta t$, constructing a speed profile naturally. Besides, considering a common rule in real-world driving, individual targets also include following the reference path indicated by the road structure (see the dashed central line in Fig. 2). To depict the points above, we write individual rewards $R_I$ as

$$R_I(\boldsymbol{x}_i) = \tau_i(\boldsymbol{x}_i^N) - \alpha \sum_{n=1}^{N} \|\boldsymbol{r}(\boldsymbol{x}_i^n) - \boldsymbol{x}_i^n\| \quad (1)$$

where $\boldsymbol{r}(\boldsymbol{x}_i^n)$ represents the on-reference point closest to $\boldsymbol{x}_i^n$; $\tau_i(\boldsymbol{x}_i^N)$ denotes the longitudinal distance from the starting position $\boldsymbol{r}(\boldsymbol{x}_i^0)$ to the end position $\boldsymbol{r}(\boldsymbol{x}_i^N)$ along the reference path; $\|\cdot\|$ is Euclidean norm; $\alpha \geq 0$ is a free parameter. In Eq. (1), the first term of the right-hand shows the desire for advancing driving progress, and the second term indicates the avoidance of path deviation. Note that to avoid the excessive sensitivity of agents' planning results to the reference path, we dynamically adjust the reference path according to agents' motion at every time step. As shown in Fig. 3, the basic idea is to generate a time-varying spline that connects the agent and the exit of the intersection for each agent. The starting and ending points of the spline are tangent to the velocity direction of the agent and the central line of the road connecting to the intersection, respectively.

*Group reward.* Due to the existence of space-sharing in interactive driving events, potential collisions with interacting agents need to be tackled. Nevertheless, simply taking collision avoidance as a target to pursue may sometimes bring near-collision solutions, which are also unfavorable. Instead, we replace the reward of avoiding a collision with the reward of mitigating the strength of conflict, decomposing the one-shot reward into each time step within the planning horizon. The strength of conflict could be measured with respect to the minimal distance between the trajectories of involved agents. Thus, to motivate actions that mitigate the conflict state, the group rewards $R_G$ is obtained by minimizing and delaying the conflict

$$R_G(\boldsymbol{x}_i, \boldsymbol{x}_{-i}) = (N - n_m + 1) \cdot \|\boldsymbol{x}_i^{n_m} - \boldsymbol{x}_{-i}^{n_m}\|^2 \quad (2)$$

where $n_m$ is the time step when the two agents are closest in distance; then the term $(N - n_m + 1)$ and $\|\boldsymbol{x}_i^{n_m} - \boldsymbol{x}_{-i}^{n_m}\|^2$ depict the temporal and spatial emergency of the conflict, respectively.

*Social preference.* We treat individual and group rewards as two parts of the motivation that form an agent's utility and guide its motion-planning process. One could either ignore the interaction and solely pursue the individual rewards, or harness the interaction and gain group rewards by cooperating with other vehicles, or, as is most frequently observed in real-world driving, take the combination of both [33]. The balance between self and group interests comes to be individualized and can be quantified with a weighted sum of two types of rewards. Mathematically, we denote the utility $U$ of an agent as

$$U = \cos(\theta) \cdot Z(R_I) + \sin(\theta) \cdot Z(R_G) \quad (3)$$

where $\theta \in [-\frac{\pi}{2}, \frac{\pi}{2}]$ is the Interaction Preference Value (IPV); $Z(\cdot)$ is the z-score normalizing function. We use IPV $\theta$ to regulate the reward weights which are symbolized as $\cos(\theta)$ and $\sin(\theta)$, measuring the social preference of individuals. Thus, IPV is a property that governs an agent's relative preference for individual rewards over group rewards. One could readily verify that the expression of utility in Eq. (3) can represent various possible relative preferences for individual and group rewards. As shown in Fig. 4, a higher value of $\sin(\theta)$ represents greater *cooperativeness* and willingness to engage in the group task, whereas a higher value of $\cos(\theta)$ means *selfishness* which indicates more attention to unilateral ego interests. In addition, a negative value of $\sin(\theta)$ indicates *competitiveness* which could trigger actions that may impede others. We assume that all drivers share the same utility components (individual and group rewards) but differ in IPV, and the IPV of a specific driver may vary over time according to real-time driving scenarios. Except for IPV, other free parameters in the utility function are calibrated on the dataset that is also applied in the following real-world driving data analysis, using the inverse reinforcement learning (IRL) approach reported in [32].

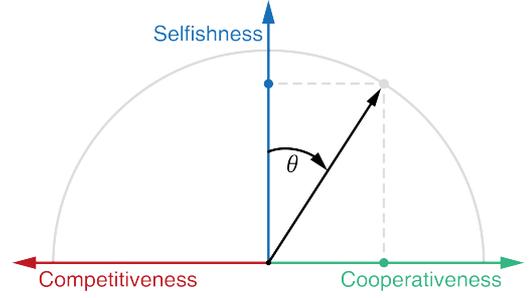

Fig. 4. Quantifying social preference with IPV.

*2) Action-dependence*

Provided that the interacting agents are motivated by maximizing their utility when planning their motion, we could then formulate agent $i$'s motion-planning problem as:

$$\boldsymbol{x}_i^* = \arg\max_{\boldsymbol{x}_i \in \mathbb{P}_i} U_i(\boldsymbol{x}_i, \boldsymbol{x}_{-i}, \theta_i) \quad (4)$$

where $\mathbb{P}_i$ is the feasible solution set of agent $i$. $\mathbb{P}_i$ is constrained by the discrete-time transition function of the vehicle dynamics $\boldsymbol{x}_i^n = \mathcal{D}(\boldsymbol{x}_i^{n-1}, \boldsymbol{u}_i^n)$ where $\boldsymbol{u}_i^n \leq \overline{u} \in \mathbb{R}^+$ is a constant control input during the time interval $t \in [(n-1)\delta t, n\delta t]$, $\overline{u}$ represents the control limit, and $\mathcal{D}(\cdot)$ is the transition function of a bicycle model. $\mathbb{P}_i$ is also limited by the inequality constraints $\|\boldsymbol{r}(\boldsymbol{x}_i^n) - \boldsymbol{x}_i^n\| \leq (w_{lane} - w_{veh})/2$ that prevent the vehicle from leaving the lane where $w_{lane}$ and $w_{veh}$ are the width of the lane and the vehicle respectively.

Because of the group target, the utility $U_i(\boldsymbol{x}_i, \boldsymbol{x}_{-i})$ jointly depends on the actions of the subjective agent and its interacting counterpart. Therefore, the problem described by Eq. (4) naturally bridges the action dependence between two interacting agents. As two agents cannot communicate with each other, the subjective agent is not supposed to directly obtain its interacting counterpart's plan, namely, $\boldsymbol{x}_{-i}$ and, this makes the motion-planning problem still unsolvable. A common solution is to give a prediction of the future motion of the interacting agent based on its current motion. However, as the motion of the interacting agent would be continuously influenced by the subjective agent's action, prediction-based methods may suffer from interaction-originated uncertainty [34]. To tackle this issue, we adopted a game-theoretic method proposed in [27] to allow the subjective agent to model its interacting counterpart as a rational game player and thus take a guess on its motion.

*Game-theoretic action-dependence formulation.* During the driving interaction, a piece of common knowledge is that every agent follows a basic strategy that one would benefit itself (by maximizing its utility) under the awareness of the interplay. Therefore, from the view of the subjective agent, we could then formulate the strategy of the interacting counterpart as:

$$x^*_{-i}(x_i): \mathbb{P}_i \to \mathbb{P}_{-i}$$

where

$$x^*_{-i}(x_i) = \arg\max_{x_{-i}} U_{-i}(x_{-i}, x_i, \hat{\theta}_{-i}) \quad (5)$$

$$s.t. \begin{vmatrix} x^n_{-i} = \mathcal{D}(x^{n-1}_{-i}, u^n_{-i}) \\ g(x^n_{-i}) \le 0 \end{vmatrix}$$

where $g(x^n_{-i}) = [r(x^n_{-i}) - x^n_{-i}]^2 - [\frac{(w_{lane}-w_{veh})}{2}]^2 \le 0$ is the lane deviation constraint. Eq. (5) provides a *best response map* with which the subjective agent could take a guess on its interacting counterpart's plan given the ego plan $x_i$ and interacting counterpart's IPV $\theta_{-i}$. As the IPV is a hidden pattern behind the driving behavior, the subjective agent needs to estimate $\theta_{-i}$ through the observation of interacting counterpart's driving behavior. An estimation method for this purpose will be presented in the ensuing subsection, and here we use $\hat{\theta}_{-i}$ to denote the estimated value to complete Eq. (5).

Taking the interacting counterpart's *best reply map* into account by substituting Eq. (5) into the subjective agent's planning problem (4), we obtain:

$$x^*_i = \arg\max_{x_i \in \mathbb{P}_i} U_i(x_i, \hat{x}_{-i}(x_i), \theta_i) \quad (6a)$$

$$s.t. \begin{vmatrix} \hat{x}_{-i}(x_i) = \arg\max_{x_{-i}} U_{-i}(x_{-i}, x_i, \hat{\theta}_{-i}) & (6b) \\ s.t. \begin{vmatrix} x^n_{-i} = \mathcal{D}(x^{n-1}_{-i}, u^n_{-i}) \\ g(x^n_{-i}) \le 0 & (6c) \end{vmatrix} \end{vmatrix}$$

where $\hat{x}_{-i}$ is the subjective agent's guess on its interacting counterpart's plan. In this bi-level optimization problem, the upper-level problem (6a) describes the motion optimization based on the estimation of the interacting counterpart's trajectory, and the lower-level problem (6b) the *best reply map* of the interacting counterpart. Eqs. (6a-b) formulates the action-dependence between interacting agents, however, the form of bi-level optimization makes it mathematically difficult to solve. To tackle this issue, we apply Karush-Kuhn-Tucker (KKT) optimality conditions [35] to the lower-level optimization and reformulate the planning problem into its single-level local equivalent:

$$x^*_i = \arg\max_{x_i, x_{-i}} U_i(x_i, x_{-i}, \theta_i) + U_{-i}(x_{-i}, x_i, \hat{\theta}_{-i}) \quad (7)$$

$$s.t. \begin{vmatrix} \nabla_{x_{-i}} U_{-i}(x_{-i}, x_i, \hat{\theta}_{-i}) + \sum_{n=1}^{N} k_n \nabla_{x^n_{-i}} g(x^n_{-i}) = 0 \\ \forall n \in \{1,2,\dots,N\}, \begin{vmatrix} x^n_i = \mathcal{D}(x^{n-1}_i, u^n_i) \\ g(x^n_i) \le 0 \\ x^n_{-i} = \mathcal{D}(x^{n-1}_{-i}, u^n_{-i}) \\ k_n g(x^n_{-i}) = 0 \\ k_n \le 0 \end{vmatrix} \end{vmatrix}$$

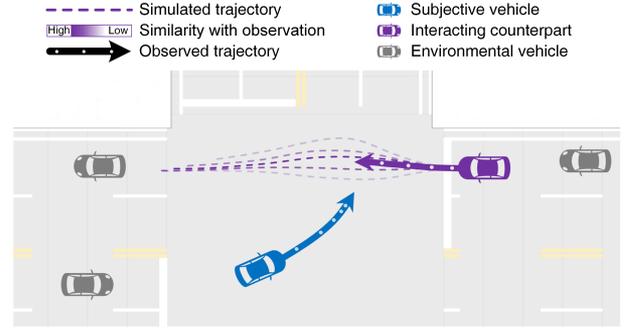

Fig. 5. Estimating IPV expression from observed trajectory. Simulated trajectories are generated under varying settings of interacting agent's IPV. The lower transparency of simulated trajectories represents a higher similarity between observed and simulated trajectories.

where $k_n$ is the Lagrange multipliers associated with the inequality constraints (6c). The utility of the interacting counterpart is added to the objective function in Eq. (7) as a trick to ensure a maximum of the primal lower-level optimization [32].

Note that in the motion-planning formulation, we avoided including any hard safety constraints due to the following considerations. First, to make interpretation of collisions possible. The main purpose of our modeling work is to interpret the driving interactions, including the potential failed cases, the results of which are collisions. Second, to model human drivers with some specific kinds of IPV. As collision avoidance is a group task, the subjective agent with ultra-selfish social preference could indeed overlook safety issues, especially when having the priority of access, leaving the group task to its interacting counterpart.

*3) Information gathering*

IPV is an individualized characteristic that guides the behaviors of a single agent. This means one could conversely deduce the IPV of a particular agent from its behaviors. With such intuition, we propose a virtual-game-based method for IPV estimation.

Firstly, the subjective agent generates hypotheses on its interacting counterpart's behavior mode. The LT agent is unacquainted with the ST agent at the beginning of the interaction event, which means no individualized prior knowledge about the ST agent can be obtained. However, the LT agent knows the ST agent's behaviors are governed by an IPV $\theta_{-i}$ within a specific range. Therefore, we generate a collection of $K$ hypotheses $\{f_{\theta_1}, \dots, f_{\theta_K}\}$ by sampling IPV $\theta_k \in \{\theta_1, \dots, \theta_K\}$ from a uniform distribution $\mathcal{U}(-\frac{\pi}{2}, \frac{\pi}{2})$. Note that aside from the uniform distribution used in this work, assuming that IPV conforms to other distributions (such as the normal distribution) with properly calibrated parameters does not affect the feasibility of the estimation process. Then, each hypothesis $f_{\theta_k}$ parameterized by $\theta_k$ represent a unique behavior mode of the ST agent. The hypotheses are applied to construct virtual ST agents in the simulation. By solving the game in simulations where the LT agent interacts with different kinds of virtual ST agents, we get a prevision on what the actual ST agent's trajectories might be like under certain kinds of IPV (Fig. 5).

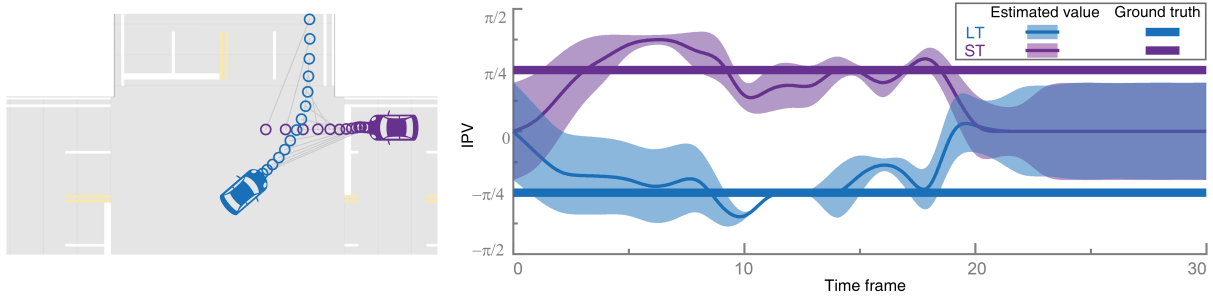

Fig. 6. Dynamic IPV estimation in a simulation case. The thick solid line means the IPV ground turth that motivates the behavior of simulated vehicles, while the thin solid curve and the shade are the IPV estimated from observed trajectories and the confidence bound respectively.

Secondly, the subjective agent compares hypotheses with the observed one. During the real-time interaction, the LT agent could continuously observe the ST agent's actual track $\bar{x}$. By comparing the actual track $\bar{x}$ with the tracks generated by each hypothesis $x_k$, we quantify each hypothesis's ability to explain the ST agent's behaviors:

$$l_k(\theta_k, \theta_{-i}) \propto p(x_k|\theta_{-i}) \propto \mathcal{N}(x_k|\bar{x}(\theta_{-i}), \sigma^2) \quad (8)$$

where $l_k(\theta_k, \theta_{-i})$ is the likelihood function that describes the similarity between two behavior modes governed by $\theta_k$ and $\theta_{-i}$; $\sigma^2$ the variance. Measuring similarity with Eq. (8) follows a straightforward assumption that the higher similarity in IPV, the higher possibility of yielding the same track. A larger $l_k$ means the greater ability to account for the ST agent's behavior. Therefore, $l_k$ is later used as the indicator for hypothesis reliability.

Thirdly, the subjective agent assembles an IPV estimator. With the awareness of the reliability of each hypothesis, we can assemble an IPV estimator with a weighted sum:

$$\hat{\theta}_{-i} = \sum_k w_k \theta_k$$

$$\sigma^2_{\hat{\theta}_{-i}} = \sqrt{\sum_k \left[w_k(\theta_k - \hat{\theta}_{-i})^2\right]}$$

where $\hat{\theta}_{-i}$ is the estimated IPV, $\sigma^2_{\hat{\theta}_{-i}}$ the uncertainty and $w_k = l_k/\sum_k l_k$ the weight of each hypothesis. By applying the estimated IPV of the ST agent in solving the game, the LT agent sets a self-fulfilling belief on the ST agent's individualized behavior mode, which is governed by $\hat{\theta}_{-i}$.

Note that as the only input needed is the observed trajectories, the IPV estimation process could be conducted from a third-party view that is independent of any interacting agents. Fig. 6 shows a simulation experiment of the driving interaction between a competitive LT vehicle ($\theta = -\pi/4$) and a cooperative ST vehicle ($\theta = \pi/4$). The estimation of both vehicles' IPV was initialized with the value of 0 and then updated at each frame using observed trajectories. Soon after, the estimated IPV of two vehicles started to fluctuate slightly around the ground truth. As social preference estimation is only meaningful in group events, the estimated value gradually vanished after the interaction finished (when at least one vehicle crossed the conflict point).

## IV. SOCIAL PREFERENCE IN HUMAN DRIVING INTERACTION

In this section, we analyze the social preference of human drivers using VGIM to provide evidence-based knowledge for humanlike AV design. Real-world driving interactions recorded in a typical two-phase intersection were applied for this purpose.

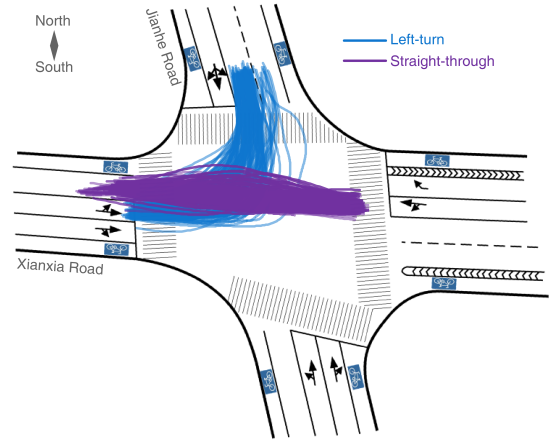

Fig. 7. Traffic trajectories in Jianhe-Xianxia intersection.

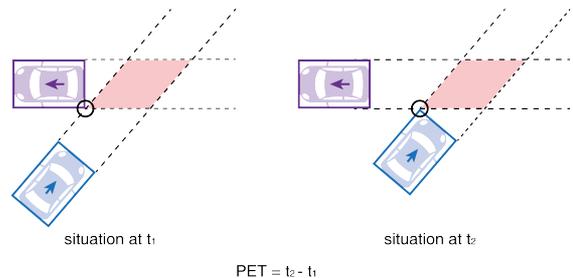

Fig. 8. Definition of Post-Encroachment Time (PET).

### A. Data collection

Interactive driving data are obtained from Jianhe-Xianxia intersection, a typical two-phase intersection in Shanghai, China. LT vehicles interact with ST vehicles frequently in this intersection and trajectories vary significantly (Fig. 7).

We set a high-definition camera on a building at the northeast corner of the intersection and recorded traffic operations during rush hours (from 4:00 pm to 5:40 pm). We extracted the temporal-spatial position of vehicles from video records with up-to-date video process assistant software [36]. Data were recorded and filtered using a Kalman Filter afterward. Overall, 210 LT trajectories and 801 ST trajectories are extracted. We then manually identified cases where LT vehicles were not affected by non-motorized road users and interacted only with a column of ST vehicles. By assuming that for any specific time point the LT vehicle interacts only with the leading ST vehicle that is approaching the conflict point, we obtained 613 paired one-on-one interaction cases.

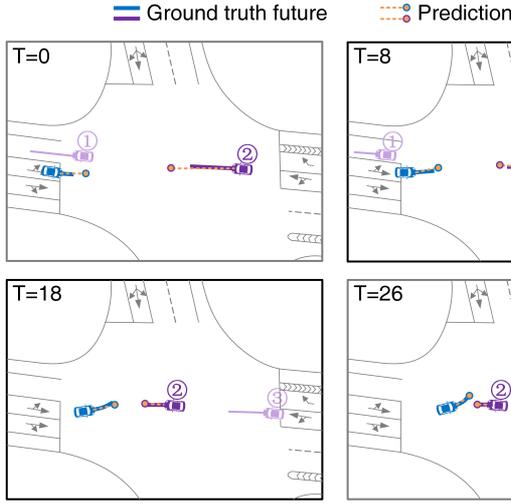
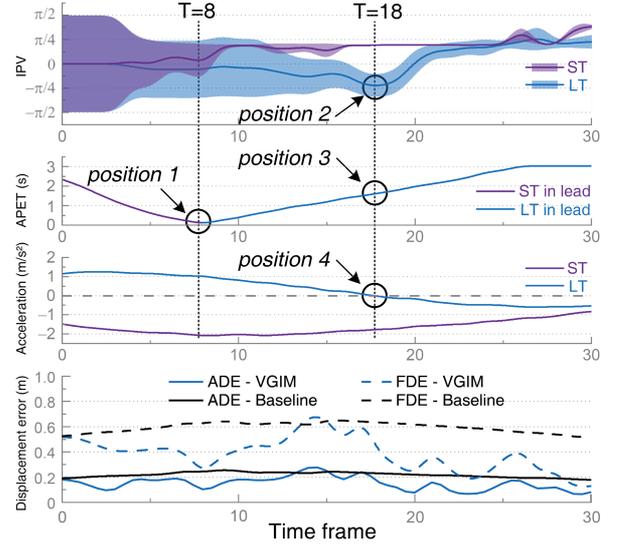

Fig. 9. Dynamic IPV estimation in a real-world left-turning driving interaction case.

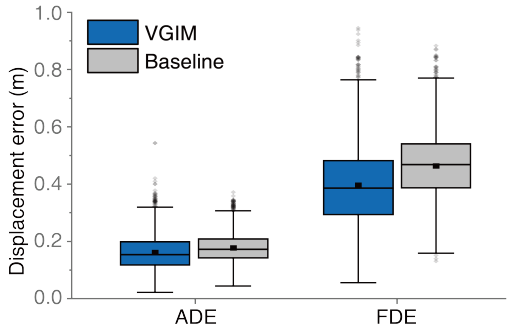

Fig. 10. Trajectory prediction error with (VGIM) and without (Baseline) IPV estimation.

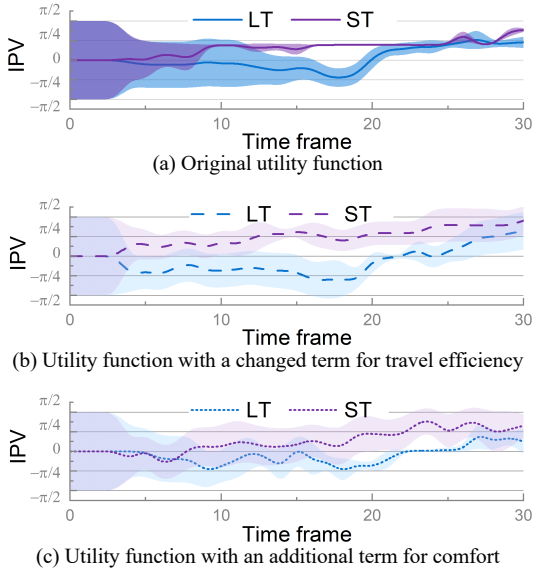

Fig. 11. IPV estimation results generated under different utility functions.

We quantified the conflict degree of each case by the anticipated post-encroachment time (APET), which is a widely-accepted indicator for measuring proximity between agents [37]. As shown in Fig. 8, PET is defined as the time gap between the moment that the first agent leaves the path of the second and the moment that the second reaches the path of the first. We calculate the APET as a real-time extension of PET at each time step using a constant speed assumption. Thus, a lower APET value means a more intense conflict state and a value of zero means a potential collision.

### B. Dynamic IPV identification

Human drivers sometimes snatch the right of way when receiving certain hints from interacting counterparts who possess a higher priority of passing. Fig. 9 shows the dynamic IPV in a case of such take-when-promised coordination conducted by human drivers in the unprotected left-turning scenario. The LT vehicle reaches the intersection and seeks a gap in the oncoming vehicle stream to cross the intersection. The LT vehicle gave up the first gap and started to interact with #2 ST vehicle at T=0. As indicated by the APET, the ST vehicle had the temporal advantage in crossing the intersection before the LT vehicle. However, the ST vehicle kept decelerating and showing an obvious social cue of letting the LT vehicle pass first. ST vehicle's behaviors are identified to be cooperative from then on. At the same time, the LT vehicle kept accelerating and decreased the APET to zero at around T=8 (*position 1*), identified to be competitive. After that, the LT vehicle took the temporal advantage and still kept, although slighter, acceleration. At T=18, the APET came to a relatively safe value (*position 3*). The LT vehicle started to decelerate afterward (*position 4*), gradually becoming cooperative (*position 2*). The ST vehicle reached the conflict point at T=26 and finished the interaction with an APET of around 3 seconds. During the interaction process, with the gradual understanding of interacting agents' social preferences, VGIM's prediction of the interaction evolvement gets more precise over time, measured by displacement error. For comparison, we give the prediction performance of a baseline by leaving out the IPV estimation process from the VGIM. At every time step, both LT and ST agents' trajectories are predicted within a 1.2-second horizon by VGIM and the baseline model. The prediction error is measured by average displacement error (ADE) and final displacement error (FDE). As shown in the right side of Fig. 9, VGIM gains obvious merits from the IPV estimation and performs better prediction indicated by smaller ADE and FDE. Form a dataset view, by applying IPV estimation, VGIM outperforms the baseline model by 9.0% on mean ADE and by 14.5% on mean FDE (Fig. 10).

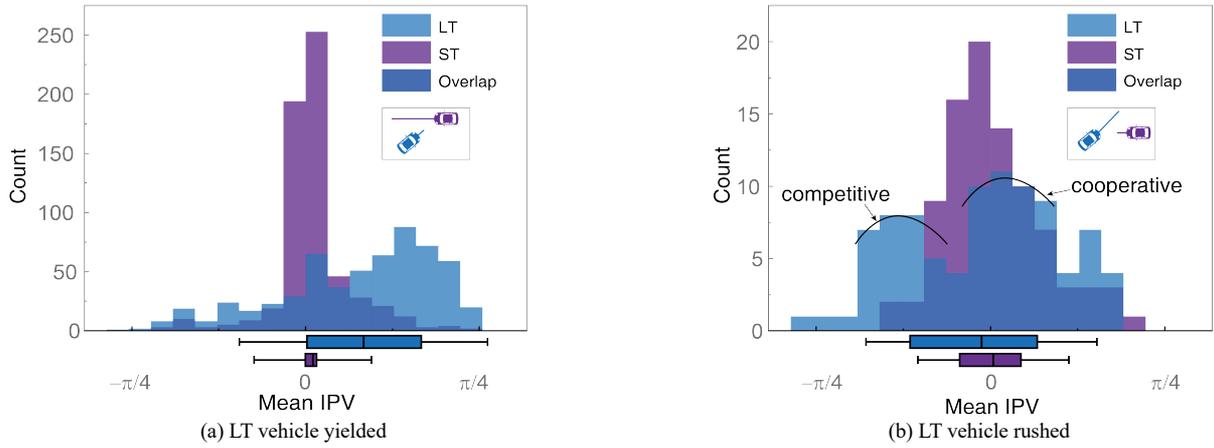

Fig. 12. Distribution of human drivers' IPV in unprotected left-turn scenario.

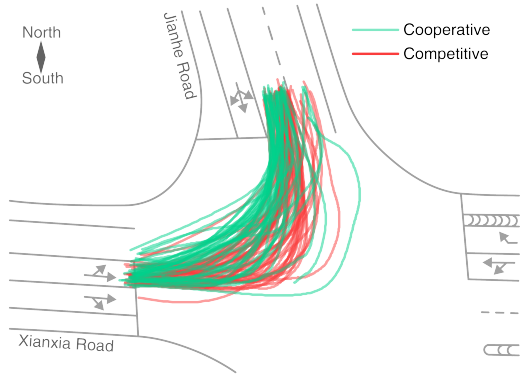

Fig. 13. Left-turn trajectories divided by IPV.

We emphasize that although the IPV estimation process is conducted based on a predesigned and calibrated utility function, the estimation results principally rely on the observed interaction trajectories rather than the utility function itself. In accordance with Eq. (8), the more adeptly the utility function captures human drivers' actual motivations in driving, the more likely we are to generate virtual agents yielding human-like trajectories, thereby producing more reliable estimates. For simplicity and generality, we presented in this work a utility function combining only the most essential terms (efficiency, traffic rule obeyance, collision avoidance) that account for generic driving behaviors and calibrated the static parameters of the utility function, specifically all the other parameters except the IPV, utilizing inverse reinforcement learning (IRL). Admittedly, the selection of particular utility terms remains empirical and somewhat subjective. To further alleviate the subjectivity of the IPV estimation process, we re-performed IPV estimation for the identical interaction event shown in Fig. 9 using distinct utility functions, modified from the original one by adding or swapping some utility terms. The utility functions for comparison were also calibrated via IRL on the same dataset. In Fig. 11, we presented the IPV estimation results employing the original utility function in Fig. 11(a) and that employing two modified utility functions in Fig. 11(b-c). In Fig. 11(b), we altered the utility term for travel efficiency into $R_{efficiency} = -\frac{1}{T}\sum_t \frac{|v_t - v_{limit}|}{v_{limit}}$ which is another well-performing form of travel efficiency suggested in [38]. In Fig. 11(c), we added a term for comfort, represented by a negatively weighted mean longitudinal jerk $R_{comfort} = -\frac{1}{T}\sum_t j_t$. Under two supplementary settings, IPV estimation results of the two modified utility functions are consistent with that of the original utility function in terms of capturing the trend of social preference expression. Despite variations in precision, the IPV results estimated by two additional utility functions still firmly support the conclusions we obtained from Fig. 9.

### C. IPV distribution in the human driver group

We estimate the dynamic IPV of agents in each left-turn case and calculate the mean IPV value to measure their average social preference. Fig. 12 shows the distributions of mean IPV in cases where LT vehicles yielded and rushed, respectively.

As shown in Fig. 12(a), the ST vehicles express near-selfish social preferences when holding the right of way, while LT vehicles tend to behave more cooperatively. In cases where the LT vehicle rushed (Fig. 12(b)), the IPV of ST vehicles varied more, and LT vehicles' IPV expression comes into two categories. Statistical results are listed in TABLE I.

TABLE I
STATISTICAL RESULTS OF ESTIMATED IPV

|  |  | LT rush cooperatively | LT rush competitively | LT yield |
|---|---|---|---|---|
| Mean IPV | LT | 0.27 | −0.34 | 0.19 |
|  | ST | 0.06 | −0.03 | 0.02 |
| Initial APET (s) |  | 2.77 | 3.51 | −5.63 |
| Initial TTCP (s) | LT | 3.07 | 4.81 | 7.44 |
|  | ST | 5.84 | 8.32 | 1.81 |

(Note: TTCP is short for time to conflict point)

We observed from TABLE I that LT vehicles choose to rush mainly when they have a temporal advantage indicated by a smaller initial Time to Conflict Point (TTCP) under constant speed assumption and show cooperativeness when the temporal advantage is weaker (indicated by smaller initial APET). For LT-rushed cases, we divided trajectories into cooperative ($\theta > 0$) and competitive ($\theta < 0$) according to the average IPV value (Fig. 13). The cooperative left-turn trajectories are featured by smaller average curvature compared to their competitive left-turn counterparts, whereas the cooperative straight-through trajectories mainly show lower speeds than the competitive ones (Fig. 14). Both LT and ST vehicles' way of showing cooperativeness echoes the motivation of pursuing the group reward defined in Eq. (2). The cooperativeness of LT vehicles is manifested by taking a straighter turn which helps move the conflict point in a direction that is spatially away from their interacting ST vehicle. By moving slower, the cooperative ST vehicles delay the time point of invading their interacting LT vehicles' path, pushing away the conflict point on the time scale.

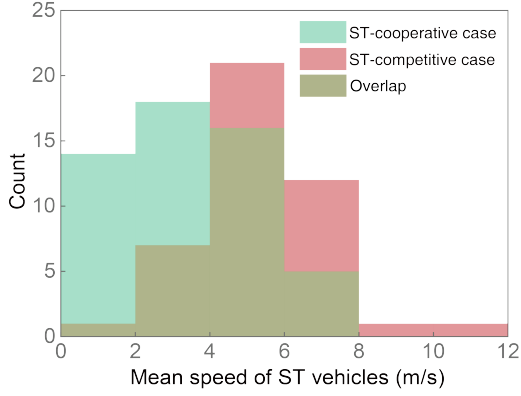

Fig. 14. Mean speed of straight-through trajectories in LT-rushed cases.

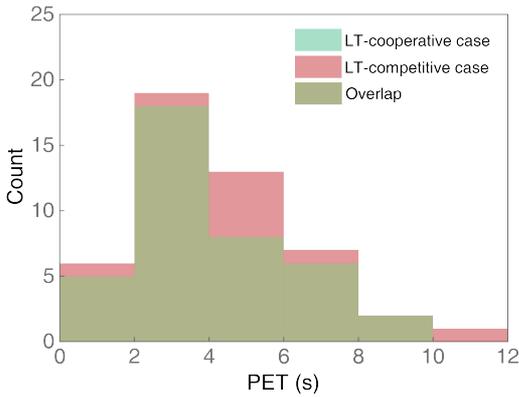

Fig. 15. PET distribution of LT-rushed cases.

## V. EXPERIMENTS AND DISCUSSION

The social preference of a single driver varies during an interaction process and differs among individuals. On the other hand, the human driver group shows stable patterns in average social preference when conducting specific driving tasks. In this subsection, we further discuss features of human drivers' social preference expression in driving interaction to provide insights for socially preferable AV design.

### A. Socially compliant IPV expression

It is a commonly accepted rule to act cooperatively when driving with lower priority in a shared space. Results of IPV analysis on LT vehicles echo this rule well when the ST vehicle is asserting the right of way (Fig. 12(a)). However, as exemplified by the case in Fig. 9, human drivers sometimes share the right of way and change the order suggested by regulations. We observed LT vehicles' stable competitiveness in a major part of LT-rushed cases (Fig. 12(b)). Wilcoxon–Mann–Whitney test found no significant differences ($p > 0.2$) between the PET of LT-competitive cases and that of the LT-cooperative cases (Fig. 15). In other words, competitive LT trajectories do not necessarily impede safety in human driving interactions. This could be partially explained by the fact that in LT-rushed cases human drivers tend to turn left competitively only when the initial state is relatively safer (TABLE I). The larger initial APET of competitive LT-rushed cases withstands hypothesis testing with statistical significance ($p = 0.07$). Thus, although competitive actions consume some redundancy for safety, the process as a whole is not necessarily more dangerous.

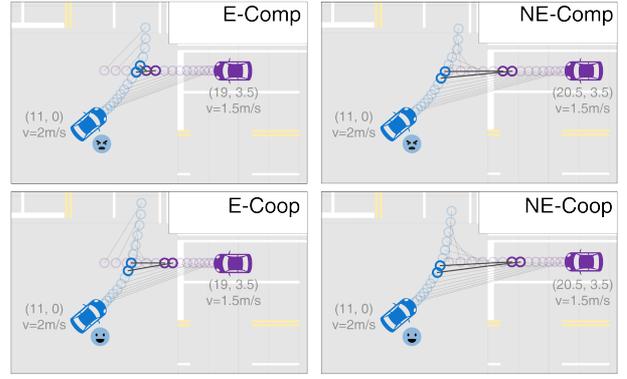

Fig. 16. Trajectories of simulation experiments.

To investigate whether the competitive IPV expression in human strategy is necessary or beneficial, we conducted controlled simulation experiments with VGIM to give counterfactual visions. The initial scenarios are designed to be emergent (**E**) and non-emergent (**NE**), and the IPV of LT vehicles is set to be competitive (**Comp**) and cooperative (**Coop**), composing 4 cases (**E-Comp**, **E-Coop**, **NE-Comp**, **NE-Coop**). The IPV of ST vehicles mirrors the average of that in human driving data. Detailed experiment settings are listed in TABLE II.

TABLE II
SETTINGS OF SIMULATION EXPERIMENTS

|  | IPV | | Initial APET (s) |
|---|---|---|---|
|  | LT | ST | (LT is in lead) |
| **E-Comp** | $-\pi/8$ | 0.06 | 2.5 |
| **E-Coop** | $\pi/8$ | 0.06 | 2.5 |
| **NE-Comp** | $-\pi/8$ | $-0.03$ | 3.5 |
| **NE-Coop** | $\pi/8$ | $-0.03$ | 3.5 |

Fig. 16 shows the results of the simulation experiments. A collision event is observed in the E-Comp case, which means turning left competitively in emergent scenarios could be perilous. In contrast, the E-Coop case stabilized at a collision-free level soon after the beginning of the interaction. Two NE cases evolved without collisions. Compared with the NE-Coop case, the NE-Comp case evolved with similar travel progress but a shorter gap time.

The events we observed in real-world driving data are likely to be the E-Coop and NE-Comp cases. According to the simulation results, it is reasonable to behave cooperatively in emergent scenarios to ensure safety. However, the NE-Comp event seems to be an irrational choice as it sacrifices part of the safety for almost nothing. We believe this is where social customs play a role. Markkula et al. [9] argued that human drivers sometimes take actions to show ego intention to others. LT vehicles' competitive actions in non-emergent scenarios could be interpreted as a strategy to make it clear to its interacting counterpart that the LT driver would not be yielding. From the perspective of AV design, it brings no direct merits to act competitively in non-emergent scenarios. Such behaviors are not likely to be produced by any Pareto-optimization-based planning method. But the NE-Comp events are repeatedly observed in human driving data and, therefore, can better fulfill the expectation of the interacting counterpart and other surrounding road users. It is relatively tractable to design an AV that does not actively cause collisions. But to act in a manner consistent with social customs, it further needs to avoid performing unpredictable maneuvers. Our findings prove that the driving strategy consistent with the social customs may generate actions that deviate from the reward-orientated optima. As simply acting cooperatively might not be socially acceptable, properly expressing competitiveness thus becomes a social skill for driving in human-involved traffic.

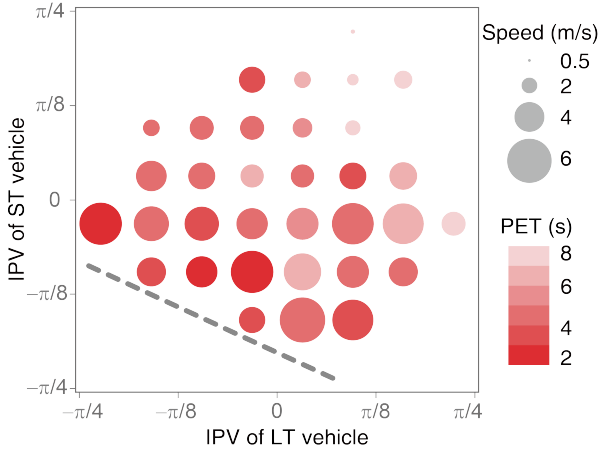

Fig. 17. Cost map of LT-vehicle-rushed interaction cases.

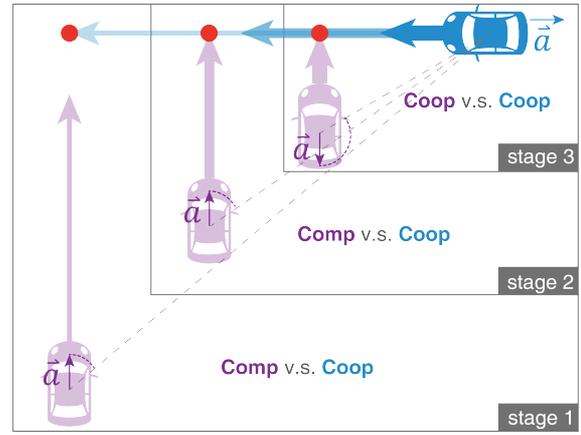

Fig. 18. Three stages in the take-when-promised driving interaction.

### B. Competitiveness in take-when-promised coordination

Interactive driving events usually happen with emerging and alleviating processes of the conflict, during which human drivers may vary their social preferences at different moments of task execution [7]. We are interested in what is the humanlike way of expressing social preferences, especially competitiveness, to navigate smoothly in such complex scenarios. We present the safety (measured by PET) and efficiency (measured by the average speed) measurements of LT-rushed cases with respect to the IPV of involved vehicles in Fig. 17. As is obtained from the human driving data, Fig. 17 can be considered the cost map of a driving strategy that follows social customs. In Fig. 17, we could find a boundary (see the dashed line) that provides a general insight into the human strategy: a pair of interacting human drivers do not show strong competitiveness simultaneously in a single interaction event.

More sophisticated patterns of social preference expression can be found in the example in Fig. 9. We divide the interaction event into three stages according to IPV expression and the scene state. Features of the three stages are presented in Fig. 18. Stage 1 is featured by the LT vehicle's competitiveness and the ST vehicle's cooperativeness. This stage serves the purpose of declaring semantic intention. ST vehicle is delivering the message that it is giving the right of way while LT vehicle accelerates to respond to ST vehicle's social cue. After the interaction in stage 1, the two vehicles confirmed each other's intention and reached an agreement on the take-when-promised coordination. Stage 2 shares the same social preference expression as in stage 1 but serves a different purpose of which is to solidify the virtual agreement by extending the gap time of passing the conflict region. With a safe gap time ensured, both two vehicles show cooperativeness at stage 3. The cooperative social preference patterns in stage 3 could be interpreted as to ensure a more controllable motion state when navigating near the conflict region.

Two points are worth noting. First, although the case shown in Fig. 9 contains most of the featured stages we observed in human driving data, it is not the most typical case as the ST vehicle in this specific case chose to give the right of way when it holds the lead of travel progress. As we found in the last subsection, in more frequently observed LT-rushed cases, the interaction begins with the LT vehicle's large temporal advantage. Second, in cases that start from stage 1, the continuous cooperativeness of the ST vehicle is an essential condition for the take-when-promised coordination. Otherwise, the interaction usually comes to the LT-yielded type.

### C. Socially compliant motion planning using VGIM

Putting together results from the statistical and case analysis above, we have found two main situations where competitiveness is expressed as part of the human strategy:

- Getting right of way from a highly cooperative interacting counterpart with higher passing priority.
- Asserting the right of way when possessing temporal advantage.

To verify the features that we found in human drivers' IPV expression, we designed a VGIM-based motion planner that captures those features with a simple rule-based IPV expressing strategy:

- Act competitively ($\theta_i = -\pi/8$) when the interacting counterpart is identified to be cooperative ($\theta_{-i} > \pi/8$), or the ego vehicle has an obvious temporal advantage

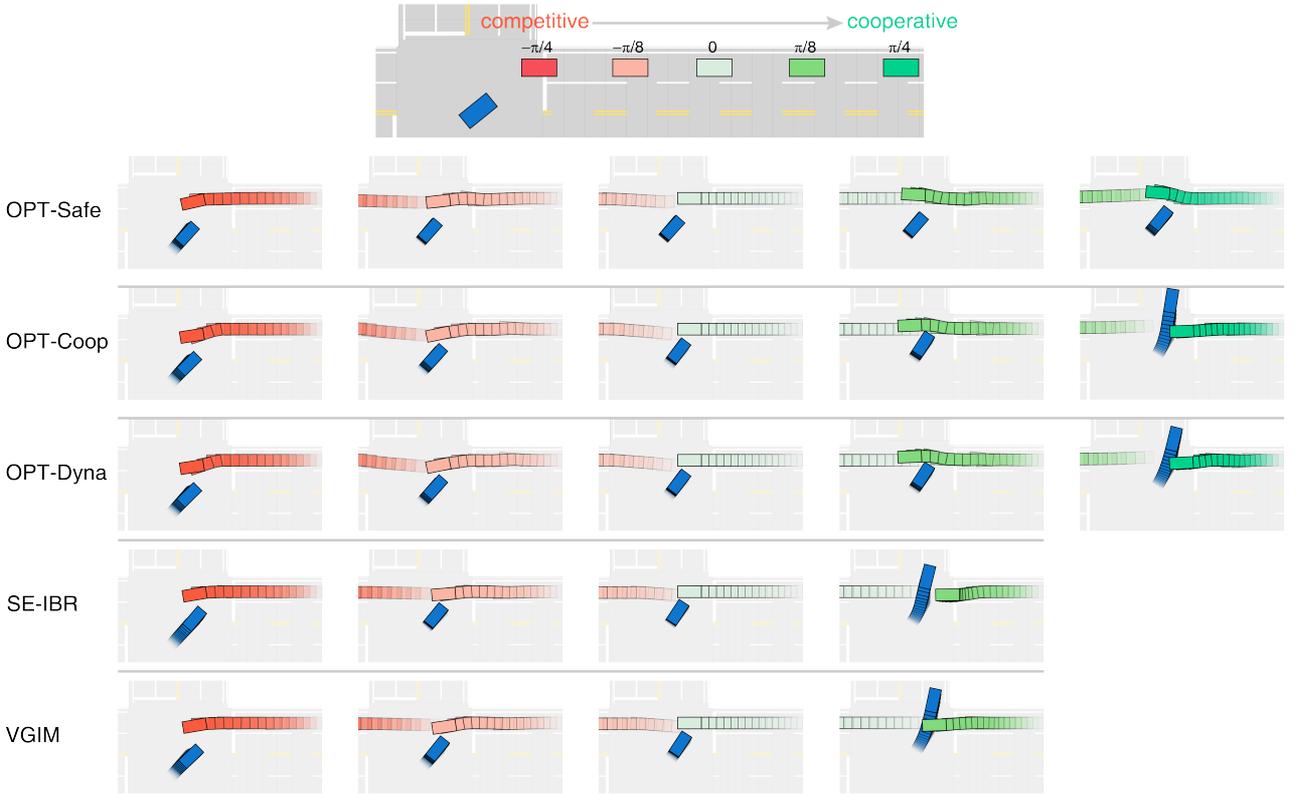

Fig. 19. Diverse behaviors in unprotected left-turn scenario.

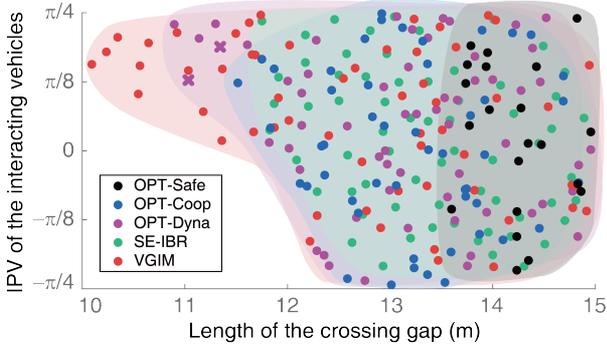

Fig. 20. Gap choosing results when interacting with endless straight-through traffic. The x shape means crashed cases.

(APET > 3 seconds).
- Act cooperatively ($\theta_i = \pi/4$) otherwise.

Given that the competitive IPV of $-\pi/8$ is approximately the average human IPV in competitive left-turn cases, and the cooperative IPV of $\pi/4$ is expressed in the most cooperative human demonstration (Fig. 12), the above rules dictate an interacting strategy that is predominantly cooperative, yet moderately competitive when required. We emphasize that, based on our observations of IPV expression in real-world interaction events, interacting counterparts with higher priority are seldom cooperative enough. This implies that, following the above rule-based strategy, the VGIM planner's action is still primarily cooperative when conducting driving tasks with lower passing priority, such as a left turn. From the perspective of equilibrium computation, a cooperative agent is more likely to yield to its interactants. Thus, without explicitly setting a rule on VGIM planner's passing order, we encourage it to yield to other road users with higher priority within the scheme of IPV expression.

We compare the VGIM planner with the following baselines: **SE-IBR** is a game theoretic planner [27] that computes the Nash equilibrium using a sensitivity-enhanced iterative best response (SE-IBR) method, being interactive but unaware of individualized sociality. For a fair comparison, we retain the original formulation of the SE-IBR planners, while applying the calibrated individual reward defined in Eq. (1) as its utility function. **OPT-Safe** is an optimal-control-based planner motivated by the calibrated individual reward defined in Eq. (1). **OPT-Coop** is an optimal-control-based planner that leverages IPV estimation to gather information on the social preference of interacting counterparts and is motivated by a utility function that equals the weights of individual and group rewards. Based on OPT-Coop, **OPT-Dyna** further applies the rule-based IPV expressing strategy to modify the utility function. We presented the details of OPT-based planners in the Appendix.

Fig. 19 shows the results in the first benchmark, where the left-turn (LT) vehicle needs to interact with a straight-through platoon. The straight-through (ST) platoon is comprised of five vehicles whose IPVs range from competitive ($\theta_{-i} = -\pi/4$) to cooperative ($\theta_{-i} = \pi/4$). The initial distance between every two ST vehicles is identical. In this benchmark, only OPT-Safe yielded to all five ST vehicles. With the awareness of IPV estimation, OPT-Coop and OPT-Dyna better predicted the interacting counterpart's motion, and both chose to rush before the most cooperative ST vehicle. VGIM and SE-IBR safely rushed before the 4th vehicle who shows slight cooperativeness. While the VGIM-involved interaction evolved generally smoothly, the SE-IBR almost compelled the interacting ST vehicle to halt. The underlying reason is that according to the rule-based IPV expressing strategy, VGIM behaves competitively when detecting the ST vehicle's cooperativeness and such early exposure of competitive preference grants the ST vehicle more time to conduct smoother responsive maneuvers.

Fig. 20 gives a comprehensive view of the five planners' decision-making strategies with the second benchmark, where the LT vehicle needs to interact with endless straight-through traffic. The ST vehicles' IPV is randomly sampled from the IPV collection obtained from human driving data. The initial distance between every two vehicles is randomly set within a range from 10 to 15 meters. With the dynamic IPV expressing strategy, VGIM accomplished more take-when-promised coordination than other planners when encountering a highly cooperative ST vehicle. As SE-IBR presumes both itself and others to be fully selfish, it shows less sensitivity to interacting counterparts' IPV. SE-IBR's trajectory solutions always converge to yielding when the LT vehicle has no temporal advantages over ST vehicles, leading to a standstill in the intersection. Consequently, if the LT vehicle fully ceased at the intersection, even a cooperative ST vehicle will not yield as no interaction actually exists, limiting SE-IBR's performance when interacting with highly cooperative ST vehicles. OPT-Safe treats every ST vehicle as a moving obstacle and crosses the intersection only when it gets a large gap. OPT-Coop and OPT-Dyna could find solutions to cross in additional situations where the gap is smaller while the interacting ST vehicle is cooperative. Compared to OPT-Coop, OPT-Dyna successfully rushed into narrower gaps but also caused some crashed cases. Collisions caused by OPT-Dyna indicate that without the awareness of interaction, the prediction process that is decoupled from the ego plan will suffer from low accuracy, hence impeding safety. This finding is in line with the results reported in [39].

To sum up, firstly, IPV estimation helps better predict the interacting counterpart's action, subsequently benefiting motion planning. This merit is observed in both VGIM and OPT-based motion planners. Secondly, compared to simply being cooperative, the dynamic IPV expressing strategy further enables coordination with cooperative interacting counterparts willing to give the right of way, but the safe implementation of dynamic IPV relies on the awareness of interaction. Note that this study tries to demonstrate that learning how human drivers express their social preferences is a simple yet promising way of emulating human-like navigation in driving interactions, but we do not claim that the VGIM-based planner as-is is the best solution. The main focus of VGIM is to capture quantified sociality patterns in human driving interactions. For the AV motion planning, we emphasize that the IPV estimation method presented in this work is independent of the planner, which means switchable IPV is a feature that is applicable to any reward/cost-based motion planning scheme.

### D. Limitations

We identify the following limitations. First, the rule-based IPV expression strategy is a preliminary trial of mimicking human driving customs. With the IPV estimation method provided by VGIM, more sophisticated patterns in socialized human driving interactions are to be explored and utilized. Second, applications of social preference adjustment for AV rely on the real-time estimation of interacting agents' IPV, while the IPV estimation method proposed in this work could be implemented in real-time with a horizon of no more than five steps due to computational complexity. For better information-gathering performance, ongoing works include a sampler-checker-based structure for faster implementation. Third, we find stable patterns of human drivers' IPV expressing strategy in the unprotected left-turn scenario, but the generalization to other interaction scenarios is unclear. The diagram of VGIM is theoretically suitable for modeling and analyzing any path-crossed driving interaction scenarios. Therefore, comparative studies are still needed.

## VI. CONCLUSION

In this work, we proposed a game-theoretic model VGIM to capture inter-agent action dependence in driving interactions. Within VGIM, the IPV serves as an indicator to quantify the individualized social preference expressed by human drivers. Through human driving data analysis, we identified the strategic patterns of social preference expression, which were subsequently validated and reproduced in simulation experiments. We conclude our findings as follows:

- Human drivers' competitive behaviors are repeatedly observed in interactive driving interactions. Although taking competitive actions in these scenarios brings no direct merits, it can be part of the socially acceptable driving strategy.
- IPV estimation enables information gathering on unacquainted interacting counterparts during driving interactions. This results in better prediction of the interacting counterpart's motion and can potentially gain merits for AV's motion planning in interactive scenarios.
- We found strategic patterns of social preference expression in human driving interactions and captured these patterns with a rule-based strategy. This strategy was then combined with VGIM and baseline motion planners to verify its effect on motion planning. The results show that humanlike IPV adjustment prompts the motion planners to play a role in take-when-promised coordination, while the awareness of interplay is essential for safely taking the right of way from the interacting counterpart.

## APPENDIX

*Details of OPT-based motion planners*

In this section, we present additional details of OPT-based baseline motion planners applied in the experiments.

### 1) Common features

In general, all three OPT-based planners follow a receding horizon planning diagram. For every time step, OPT-based planners generate a controllable state sequence $x_i^{0 \to N}$ to maximize their utility for a finite time horizon $N$ based on ego motion state and the prediction of interacting counterpart's states $\hat{x}_{-i}^{0 \to N}$:

$$x_i^* = \arg \max_{x_i \in \mathbb{P}_i(\hat{x}_{-i}^{0 \to N})} U_i\left(x_i^{0 \to N}, \hat{x}_{-i}^{0 \to N}\right)$$

Here, $x_i$ shares the same constraints of vehicle dynamics and road boundary as that depicted in the specifications of Eq. (4). Additionally, OPT-based planners are banned from applying a trajectory that invades the predicted interacting counterpart's trajectory, i.e., $\forall n \in [1, N] \cap \mathbb{Z}, \|x_i^n - \hat{x}_{-i}^n\| > \varepsilon$, where $\varepsilon$ is the minimal safe distance between two cars. Then, the OPT-

based planner will execute an action to reach the first state in the optimal state sequence $x_i^*$.

Three OPT-based planners are diverse in their utility function and the way they predict the motion of their interacting counterparts.

*2) OPT-Safe*

OPT-Safe planner cares only about individual interests and assumes its interacting counterparts also to be so. Therefore, OPT-Safe planner degrades the utility function into the individual reward function defined in Eq. (1) for both itself and its interacting counterparts.

$$x_i^* = \arg\max_{x_i \in \mathbb{P}_i(\hat{x}_{-i}^{0 \to N})} R_I(x_i^{0 \to N})$$

$$s.t. \quad \hat{x}_{-i}^{0 \to N} = \arg\max_{x_{-i}} R_I(x_{-i}^{0 \to N})$$

*3) OPT-Coop*

OPT-Coop planner equals individual and group rewards and is aware of the individualized social preference of its interacting counterpart. For each time step $t$, OPT-Coop planner updates its interacting counterpart's IPV by applying the *information-gathering* process we proposed in Section III-B-3). The estimated IPV $\theta_{-i}^t$ is substituted into Eq. (3) to generate a real-time utility function for the interacting counterpart:

$$U_{-i}^t = \cos(\theta_{-i}^t) \cdot Z(R_I) + \sin(\theta_{-i}^t) \cdot Z(R_G)$$

Then, OPT-Coop maximizes its utility function $U_i$ while staying out of the interacting counterpart's optimal trajectory that maximizes $U_{-i}^t$:

$$x_i^* = \arg\max_{x_i \in \mathbb{P}_i(\hat{x}_{-i}^{0 \to N})} U_i(x_i^{0 \to N})$$

$$s.t. \quad \begin{vmatrix} \hat{x}_{-i}^{0 \to N} = \arg\max_{x_{-i}} U_{-i}^t(x_{-i}^{0 \to N}, \hat{x}_i^{0 \to N}) \\ U_i = Z(R_I) + Z(R_G) \end{vmatrix}$$

Here, as the interacting counterpart also takes group rewards into account, it theoretically needs to predict OPT-Coop planner's motion $\hat{x}_i^{0 \to N}$. We start the prediction by applying a constant-speed-constant-steering assumption for OPT-Coop planner to generate $\hat{x}_i^{0 \to N}$.

*4) OPT-Dyna*

Compared with OPT-Coop planner, OPT-Dyna planner shares the same process of predicting interacting counterpart's motion states but differs in the utility function. OPT-Dyna planner's IPV switches between cooperative and competitive by following the rule-based IPV expressing strategy $f(x_i^t, x_{-i}^t, \theta_{-i}^t)$ we presented in Section V-C:

$$x_i^* = \arg\max_{x_i \in \mathbb{P}_i(\hat{x}_{-i}^{0 \to N})} U_i(x_i^{0 \to N})$$

$$s.t. \quad \begin{vmatrix} \hat{x}_{-i}^{0 \to N} = \arg\max_{x_{-i}} U_{-i}^t(x_{-i}^{0 \to N}, \hat{x}_i^{0 \to N}) \\ U_i = \cos(\theta_i^t) \cdot Z(R_I) + \sin(\theta_i^t) \cdot Z(R_G) \\ \theta_i^t = f(x_i^t, x_{-i}^t, \theta_{-i}^t) \end{vmatrix}$$

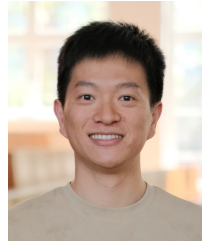

**Xiaocong Zhao** (Student Member, IEEE) received the B.S. and M.S. degree in vehicle engineering from Jiangsu University, China. He is currently working toward a Ph.D. degree in transportation engineering with the College of Transportation Engineering, Tongji University, Shanghai, China. He also researches as a joint Ph.D. student with the "Friedrich List" Faculty of Transport and Traffic Sciences, Technische Universität Dresden, Germany. His main research interest includes social behaviors in driving interaction, decision making in interactive driving scenarios, and computational game theory.

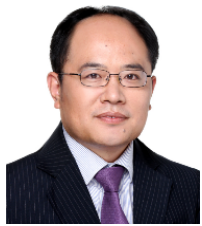

**Jian Sun** received his Ph.D. in Tongji University in 2006. Subsequently, he was at Tongji University as a Lecturer, and then promoted to the position as a Professor in 2011. He is currently a Professor with the College of Transportation Engineering and the Dean of the Department of Traffic Engineering. His main research interests include traffic flow theory, traffic simulation, connected vehicle-infrastructure system, and intelligent transportation systems.

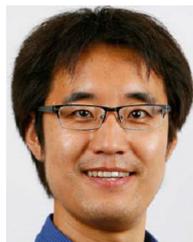

**Meng Wang** (Member, IEEE) received the B.Sc. degree from Tsinghua University in 2003, the M.Sc. degree from the Research Institute of Highway (RIOH), Ministry of Transport, in 2006, and the Ph.D. degree (Hons.) from TU Delft in 2014. He was an Assistant Professor (tenured in 2019) at the Department of Transport and Planning of TU Delft, from 2015 to 2021 and the Co-Director of the Electric and Automated Transport Laboratory (hEAT lab). From 2006 to 2009, he was an Assistant Researcher at the National ITS Center of RIOH and a Post-Doctoral Researcher at the Automotive Group, Faculty of Mechanical Engineering, TU Delft, from 2014 and 2015. He is a Full Professor (W3) and the Head of the Chair of Traffic Process Automation with the "Friedrich List" Faculty of Transport and Traffic Sciences, Technische Universität Dresden. His main research interests are traffic flow modelling and control, driver behaviour, control design, and impact assessment of connected and automated vehicles. He was a recipient of the IEEE ITS Society Best Ph.D. Dissertation Award in 2015 and the IEEE International Conference on Intelligent Transportation Systems (ITSC) Best Paper Award in 2013. He is an Associate Editor of the journal IEEE TRANSACTIONS OF INTELLIGENT TRANSPORTATION SYSTEMS, IET ITS, and Transportmetrica B and the Editorial Board Member of Transportation Research Part C.